\pdfoutput=1

\documentclass[11pt]{article}

\usepackage[preprint]{acl}

\usepackage{times}
\usepackage{latexsym}

\usepackage[T1]{fontenc}

\usepackage[utf8]{inputenc}

\usepackage{microtype}

\usepackage{inconsolata}

\usepackage{booktabs}
\usepackage{graphicx}
\usepackage{multirow}
\usepackage{makecell}
\usepackage{color}
\usepackage{framed}
\usepackage{amssymb}
\usepackage{colortbl}
\usepackage[utf8]{inputenc}
\usepackage{algorithm}
\usepackage{algpseudocode}  
\usepackage{subcaption}
\usepackage{caption}
\usepackage{tabularray}
\usepackage{xspace}
\usepackage{amsmath}
\usepackage{adjustbox}
\usepackage{makecell} 
\usepackage{soul}
\usepackage{enumitem}
\usepackage[normalem]{ulem}
\useunder{\uline}{\ul}{}

\DeclareMathOperator*{\argmin}{arg\,min}
\usepackage{listings}
\definecolor{codegreen}{rgb}{0,0.6,0}
\definecolor{codegray}{rgb}{0.5,0.5,0.5}
\definecolor{codepurple}{rgb}{0.58,0,0.82}
\definecolor{backcolour}{rgb}{0.95,0.95,0.92}

\lstdefinestyle{mystyle}{
    backgroundcolor=\color{backcolour},   
    commentstyle=\color{codegreen},
    keywordstyle=\color{magenta},
    numberstyle=\tiny\color{codegray},
    stringstyle=\color{codepurple},
    basicstyle=\ttfamily\scriptsize,
    breakatwhitespace=false,         
    breaklines=true,                 
    captionpos=b,                    
    keepspaces=true,                 
    numbers=left,                    
    numbersep=5pt,                  
    showspaces=false,                
    showstringspaces=false,
    showtabs=false,                  
    tabsize=2
}

\title{Personalized Text Generation with Contrastive Activation Steering}
\author{
Jinghao Zhang\textsuperscript{1,2}, 
Yuting Liu\textsuperscript{3}, 
Wenjie Wang\textsuperscript{4}, \\ 
{\bf 
Qiang Liu\textsuperscript{1,2}, 
Shu Wu\textsuperscript{1,2}, 
Liang Wang\textsuperscript{1,2},
Tat-Seng Chua\textsuperscript{5} } \\
\textsuperscript{1}NLPR, 
Institute of Automation, Chinese Academy of Sciences, \\
\textsuperscript{2}School of Artificial Intelligence, University of Chinese Academy of Sciences, \\
  \textsuperscript{3}Northeastern University, China,
  \textsuperscript{4}University of Science and Technology of China, \\
  \textsuperscript{5}National University of Singapore \\
  \texttt{jinghao.zhang@cripac.ia.ac.cn}, \texttt{\{qiang.liu,shu.wu,wangliang\}@nlpr.ia.ac.cn}, \\
  \texttt{yutingliu@stumail.neu.edu.cn}, 
  \texttt{wenjiewang96@gmail.com},
  \texttt{dcscts@nus.edu.sg},\\
  }

\begin{document}
\newcommand{\themodel}{StyleVector\xspace}
\newcommand{\note}[1]{{\textcolor{red}{NOTE: #1}}}
\newcommand{\red}[1]{\sethlcolor{red}\hl{#1}}
\newcommand{\green}[1]{\sethlcolor{green}\hl{#1}}
\definecolor{hlblue}{HTML}{96FFFB}

\newcommand{\blue}[1]{\sethlcolor{hlblue}\hl{#1}}

\maketitle
\begin{abstract}
Personalized text generation aims to infer users' writing style preferences from their historical texts and generate outputs that faithfully reflect these stylistic characteristics. Existing solutions primarily adopt two paradigms: retrieval-augmented generation (RAG) and parameter-efficient fine-tuning (PEFT). While these approaches have advanced the field, they suffer from two critical limitations: (1) the entanglement of content semantics and stylistic patterns in historical texts impedes accurate modeling of user-specific writing preferences; and (2) scalability challenges arising from both RAG's inference latency by retrieval operations and PEFT's parameter storage requirements for per user model. To overcome these limitations, we propose \themodel, a training-free framework that disentangles and represents personalized writing style as a vector in LLM's activation space, enabling style-steered generation during inference without requiring costly retrieval or parameter storage. Comprehensive experiments demonstrate that our framework achieves a significant 8\% relative improvement in personalized generation while reducing storage requirements by 1700 $\times$ over PEFT method. 
\end{abstract}

\section{Introduction}
Large language models (LLMs) have demonstrated unprecedented capabilities in text generation and complex reasoning through pre-training on massive corpora. However, these models still function as "one-size-fits-all" systems, optimized for average-case scenarios, and fail to adapt to individual users' unique preferences. The increasing demand for personalized AI assistants highlights the need to customize LLMs to better align with the specific preference of each user~\cite{kirk2024benefits,chen2024large,au2025personalized,cai2024large,jang2023personalized,lin2024persona,zhang2024personalization,Liu2024DynamicGO,Zhu2025InvestigatingLL}.

Personalized text generation has emerged as a critical research frontier~\cite{salemi-etal-2024-lamp,kumar2024longlamp,Alhafni2024PersonalizedTG,Chen2024UsingPT}. Consider a scenario where given an email subject $x$ and a user $u$'s historical subject-email pairs $P_u$, the system must infer the user's writing style from $P_u$ to generate stylistically consistent emails. Current approaches predominantly fall into two categories: (1) Retrieval-augmented generation (RAG) methods~\cite{zhang2023memory,salemi2024comparing,salemi2024learning}, which enhance input prompts by retrieving personalized information from $P_u$, and (2) parameter-efficient fine-tuning (PEFT) methods~\cite{salemi2024comparing,tan2024personalized,zhuang2024hydra}, which train per-user adapter modules using $P_u$. Despite their merits, these methods suffer from critical limitations: (a) The inherent entanglement of \emph{user-agnostic content semantics} and \emph{user-specific stylistic patterns} in historical data impedes accurate style inference. (b) The substantial inference latency of RAG's retrieval mechanisms and storage requirements of PEFT's per-user parameters renders these solutions impractical for real-world deployment at scale.

Recent advances in activation engineering~\cite{zou2023representation,liu2023context,rimsky-etal-2024-steering} reveal that LLMs encode features and concepts as linear directions in hidden activation space. These directional vectors can effectively steer model behavior through simple linear interventions during inference. Building on these insights, we reveal that \textit{user-specific writing styles} can similarly be represented as directional vectors in activation space. This leads to an elegant solution for personalized generation: (1) {By contrasting the hidden activations between \textit{user-authentic responses} (containing both content and style) and \textit{model-generated generic responses} (content-preserving but style-agnostic), we can derive "style vector" that contains personal stylistic signatures. (2) The derived style vector could be used to steer model generation towards user-specific writing styles through simple linear interventions during inference, without parameter updates or extensive retrieval.

To this end, we present \themodel, an efficient, training-free framework that only requires storing one vector for each user to achieve high-quality personalized text generation. As shown in Figure~\ref{fig:model}, our methodology comprises three key steps: (1) generating style-agnostic responses for historical inputs using a base LLM, (2) deriving style vectors by contrasting hidden activations between authentic user responses and generated neutral responses, and (3) steering generation during inference through linear activation interventions with the obtained style vectors.

Comprehensive evaluations on LaMP~\cite{salemi-etal-2024-lamp} and LongLaMP~\cite{kumar2024longlamp} benchmarks for short- and long-form personalization respectively demonstrate our method's effectiveness. Experimental results show that \themodel achieves 8\% relative improvement in personalization quality while reducing storage requirements by 1700$\times$ over PEFT-based methods.


Our contributions are summarized as follows:
\begin{itemize}
    \item We reveal that user-specific writing styles can be represented as linear directions in activation space through contrastive analysis between authentic user responses and style-agnostic model outputs.
    \item We propose a training-free personalized generation framework through simple linear activation interventions, requiring only $2|P_u|$ forward passes (zero back-propagation) per user and compresses personalized information into a single vector.
    \item Experiments on both short- and long-form personalization benchmarks show the effectiveness of our method, while significantly reducing storage and inference latency compared to retrieval-based and adapter-based approaches.
\end{itemize}

\section{Preliminaries}
\subsection{Problem Formulation}
Personalized text generation aims to infer the user's writing style preferences based on the text created from their history and generate outputs that align with those preferences. Formally, for each user $u$: given an input prompt $x$ specifying task requirements (e.g., an email subject), the language model $M$ generates output $\hat{y}=M(x, P_u)$ conditioned on both $x$ and the user's historical data $P_u = \{(x_i, y_i)\}_{i=1}^{|P_u|}$, where each pair $(x_i, y_i)$ represents previous interactions (e.g., subject-email pairs). The ground truth output $y$ represents the user-customized response that reflects $u$'s unique writing style (e.g., personalized email drafts).

\subsection{Base Solutions}

\paragraph{Retrieval-Augmented Generation (RAG)}
RAG-based approaches achieve personalization through context-aware retrieval. Given input $x$, the system retrieves $k$ most relevant historical responses from $P_u$ using retriever $R$, then generates personalized responses by combining retrieved documents $R(x,P_u,k)$ with the input prompt:
\begin{equation}
\hat{y} = M(x, R(x,P_u,k))
\end{equation}

\paragraph{Parameter-Efficient Fine-Tuning (PEFT)}
PEFT methods customize LLMs by training lightweight adapters (e.g., LoRA~\cite{hu2021lora}) on user-specific data while keeping base model parameters frozen~\cite{tan-etal-2024-democratizing}. For each user $u$, a distinct adapter $\theta_u$ is trained via:
\begin{equation}
\label{eq:sft}
\theta_u^* = \arg\min_{\theta} \sum_{(x_i,y_i) \in P_u} \mathcal{L}(M(x_i; \theta), y_i)
\end{equation}
where $\mathcal{L}(\cdot)$ denotes the sequence-to-sequence cross-entropy loss. During inference:
\begin{equation}
\hat{y} = M(x; \theta_u)
\end{equation}

\begin{figure*}[h]
    \centering
    \includegraphics[width=\linewidth]{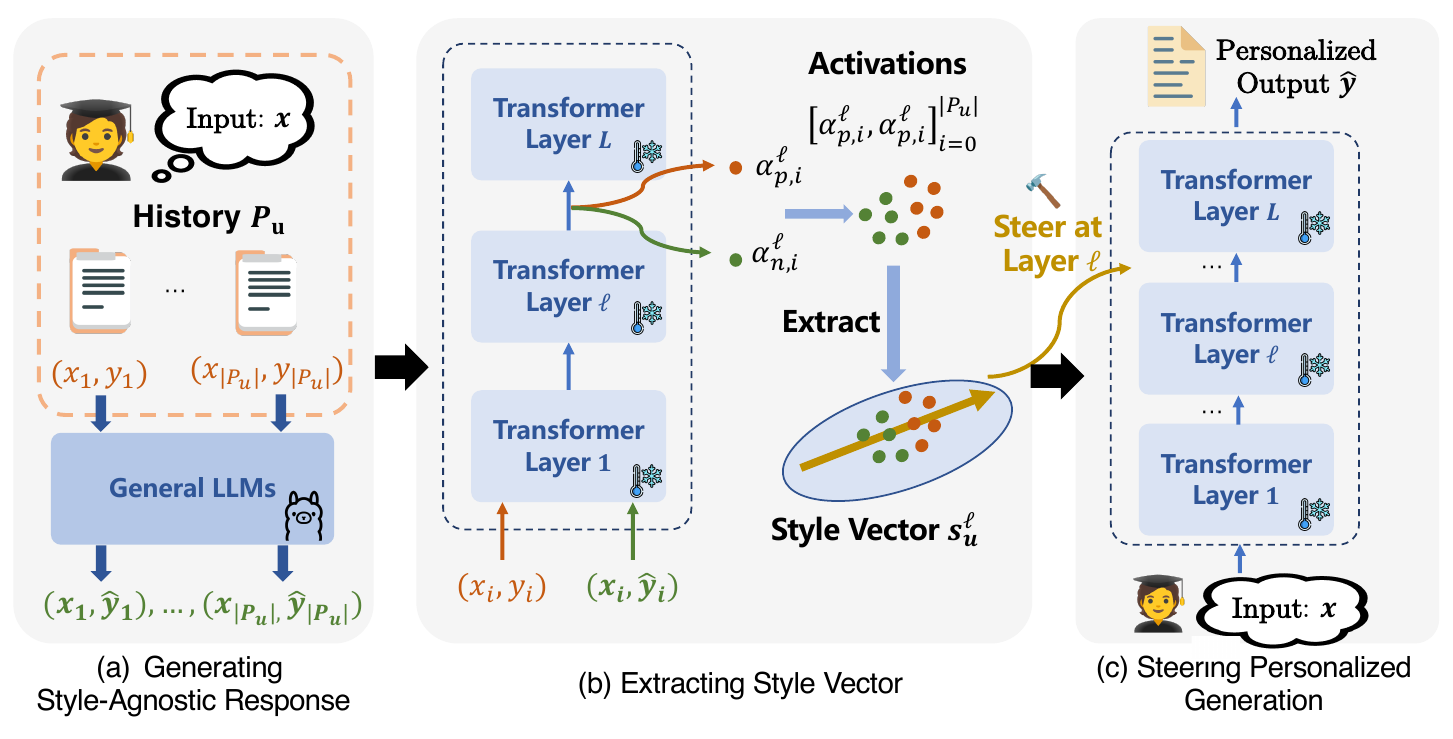} 
    \caption{The overall framework of \themodel.}
    \label{fig:model}
\end{figure*}

\subsection{Limitations of Base Solutions}  
Existing approaches face the following two fundamental constraints.

\paragraph{Entangled Style-Content Representation} Both RAG and PEFT methods process historical entries $p_i$ as monolithic units. However, each historical entry contains both the \emph{user-agnostic semantics} corresponding to the input $x_i$ and the \emph{user-specific writing style}~\cite{Fisher2024StyleRemixIA}. This entanglement impedes accurate style modeling, particularly for RAG methods that retrieve documents based on semantic matching, and the semantic-dominated retrieved contexts lead to style dilution (see Section~\ref{sec:discrete} for examples).

\begin{table}[t]
\centering
\begin{adjustbox}{max width=\linewidth}

\begin{tabular}{lccc}
\toprule
\textbf{Metric} & \textbf{RAG} & \textbf{PEFT} & \textbf{\themodel} \\
\midrule
Training Time/User & $O(|P_u|)$*  & $O(|P_u|)$ & $O(|P_u|)$* \\

Latency/Query & $O(|P_u|)$ & $O(\text{Load+Merge})$ & $O(1)$ \\

Storage/User & $O(|P_u| D)$ & $O(r D  L)$ & $O(D)$ \\
\bottomrule
\end{tabular}
\end{adjustbox}
\vspace{-0.2cm}
\begin{flushleft}
\footnotesize
* Training-free. Denotes pre-processing cost.
\end{flushleft}
\caption{System Efficiency Comparison.}
\label{tab:efficiency}
\end{table}

\paragraph{Scalability Bottlenecks}
As summarized in Table~\ref{tab:efficiency}, existing methods suffer from three critical scalability constraints:  training time, inference latency and storage requirement. Due to space constraints, we have placed the complexity analysis of the baseline in the Appendix~\ref{app:efficiency}. These compounded costs render existing methods challenging for real-world deployment at scale~\cite{salemi2024comparing}.  We also provide empirical cost comparisons in Section~\ref{sec:cost}.

\section{Method}
Our \themodel framework aims to identify a user-specific style vector through contrastive activation analysis, then steer LLM generation via targeted activation intervention. As shown in Figure~\ref{fig:model}, the process comprises three stages: (1) Style-agnostic response generation, (2) Style vector extraction through contrastive activation analysis, and (3) Activation steering during inference.

\subsection{Generating Style-Agnostic Response} 
Given a user $u$ with historical interactions $P_u = \{(x_i, y_i)\}_{i=1}^{|P_u|}$, where $x_i$ denotes an input and $y_i$ the user-authored response, we first generate style-agnostic responses $\{\hat{y}_i\}_{i=1}^{|P_u|}$ by instructing any general LLM $M_g$ with the input $x_i$:
\begin{equation}
\label{eq:general_model}
\hat{y}_i = M_g(x_i).
\end{equation}
Please note that the general LLM $M_g$ is designed to generate responses that are independent of the user's style and only related to the input semantics. It does not necessarily need to be the same as a personalized large model $M$; it can be any model, whether open-source or closed-source. We conduct experiments in Appendix~\ref{app:base_model} to show the robustness on $M_g$ of our method.

In this way, $y_i$ denotes the user-authentic content, containing both content semantics and stylistic patterns. Model-generated generic $\hat{y}_i$ only preserves content semantics related to $x_i$ but stripped of personal style. By contrasting $y_i$ and $\hat{y}_i$, we could disentangle user-specific style from user-agnostic semantics.

\subsection{Extracting Style Vector} 
We extract style vectors through contrastive analysis of hidden activations. Let $h_\ell(r) \in \mathbb{R}^d$ denote the hidden states of the last token at layer $\ell$ when processing text $r$. The positive and negative activations of history piece $i$ can be represented as:
\begin{equation}
    \label{eq:agg}
    a^\ell_{p, i} = h_\ell(x_i \oplus y_i), \quad a^\ell_{n, i} = h_\ell(x_i \oplus \hat{y}_i), 
\end{equation}
where $\oplus$ denotes concatenation the strings of input and output. Then we can obtain the user style vector by considering all history pieces:
\begin{equation}
    s_u^\ell = f([a^\ell_{p, i}, a^\ell_{n, i}]_{i=0}^{|P_u|}),
\end{equation}
where $f(\cdot)$ is an extracting function that takes all the positive and negative activations and returns a single style vector. The essence of the function \( f \) is to find a direction in the activation space that points from style-agnostic samples to user-authentic samples. There could be many possible functions, and here we discuss three strategies:

1) Mean Difference. The most straightforward approach computes the mean difference between positive and negative activations:
\begin{equation}
    s_u^\ell = \frac{1}{|P_u|} \sum_{i=1}^{|P_u|} (a^\ell_{p,i} - a^\ell_{n,i}).
\end{equation}
$s_u^\ell$ represents the average direction in the activation space that distinguishes user-specific style patterns from style-agnostic ones.

2) Logistic Regression. We can also employ logistic regression to find a direction that best separates positive and negative examples. Let $X = [a^\ell_{p,1}; ...; a^\ell_{p,|P_u|}; a^\ell_{n,1}; ...; a^\ell_{n,|P_u|}]$ be the matrix of all activations, and $y = [1,...,1,-1,...,-1]$ be the corresponding labels. The style vector is obtained by:
\begin{equation}
     w = \argmin_{w} \sum_{i} \log(1 + e^{-y_i X_i w}),
\end{equation}
where $w$ denotes the normal vector to the decision boundary. When moving in the direction of $w$, the model's predicted probability of being a positive sample will monotonically increase. We use the normalized $w$ as the style vector:
\begin{equation}
    s_u^\ell = \frac{w}{\|w\|_2},
\end{equation}

3) Principal Component Analysis. The Principal Component Analysis (PCA) approach finds the steering vector \(s_u^\ell\) by identifying the direction of maximum variance in the differences between positive and negative activations.  Let \(\Delta_i = a^\ell_{p,i} - a^\ell_{n,i}\) be the difference between the \(i\)-th pair of positive and negative activations. PCA computes the first principal component of the set \(\{\Delta_i\} \cup \{-\Delta_i\}\), which can be formulated as:

\begin{equation}
s_u^\ell = \arg\max_{v: \|v\|=1} \sum_{i=1}^{|P_u|} (\Delta_i^T v)^2.
\end{equation}
This formulation ensures that:
1. The resulting vector \(s_u^\ell\) has unit norm
2. It maximizes the projected variance of the activation differences
3. The inclusion of \(-\Delta_i\) enforces symmetry around the origin, making the solution invariant to the choice of which sample is positive or negative

The solution to this optimization problem is given by the first eigenvector of the matrix \(\sum_{i=1}^{|P_u|} (\Delta_i\Delta_i^T + (-\Delta_i)(-\Delta_i^T))\), which can be efficiently computed using Singular Value Decomposition (SVD).

\subsection{Steering Personalized Generation}
After obtaining the style vector, we can steer the model's generation by intervene the hidden states at inference time. In this work, we only consider intervene one layer $\ell$, which could be selected via validation set. Let $h_\ell(x)$ denote the hidden states at layer $\ell$ when processing input $x$. We use the most straightforward approach directly adds the scaled style vector to the hidden states of the token position $t$:
\begin{equation}
    \label{eq:steer}
    h'_\ell(x)_t = h_\ell(x)_t + \alpha s_u^\ell
\end{equation}
where $\alpha$ is a scaling factor controlling the strength of steering.  Following~\cite{rimsky-etal-2024-steering}, we intervene every token position of the generated text after the end of the initial prompt $t \geq |x|$. We also try different positions experimentally in Section \ref{sec:layer}.

\paragraph{Efficiency Analysis} For pre-processing, our method requires only $2|P_u|$ forward passes of LLMs to obtain activations and the style vector extracting is negligible when compared with the cost of LLMs.
For storage, the final style vector $s_u^\ell$ only requires $D$-dimensional vector storage. 
For additional inference latency, activation steering only introduces $D$ element-wise addition overhead. The complexity analysis is summarized in Table~\ref{tab:efficiency}.

\begin{table*}[]
\begin{adjustbox}{max width=\textwidth}
\begin{tabular}{@{}ccccccccc@{}}
\toprule
\multirow{2}{*}{\textbf{Benchmark}}                     & \multirow{2}{*}{\textbf{Metric}} & \textbf{Non-personalized} & \multicolumn{2}{c}{\textbf{RAG-based}} & \multicolumn{2}{c}{\textbf{PEFT-based}} & \multirow{2}{*}{\textbf{Ours}} & \multirow{2}{*}{\textbf{Improv.}} \\ \cmidrule(r){3-3} \cmidrule(r){4-5}  \cmidrule(r){6-7}
                                               &                         & LLaMA2           & BM25       & Contriever       & \multicolumn{1}{c}{SFT}  & DPO &                       &                          \\ \midrule
\multirow{2}{*}{LongLaMP:   Abstract Generation}    & ROUGE-L & {\ul 0.2056} & 0.2020 & 0.2035       & 0.2038       & 0.2020 & \textbf{0.2060} & 0.2\%  \\
                                                    & METEOR  & {\ul 0.2950} & 0.2911 & 0.2922       & 0.2929       & 0.2933 & \textbf{0.2973} & 0.8\%  \\
\midrule
\multirow{2}{*}{LongLaMP:   Topic Writing}          & ROUGE-L & 0.1299       & 0.1235 & 0.1256       & {\ul 0.1303} & 0.1277 & \textbf{0.1361} & 4.7\%  \\
                                                    & METEOR  & 0.1874       & 0.1782 & 0.1853       & {\ul 0.1914} & 0.1901 & \textbf{0.1949} & 4.0\%  \\
\midrule
\multirow{2}{*}{LongLaMP:   Review Generation}      & ROUGE-L & 0.1380       & 0.1388 & {\ul 0.1391} & 0.1364       & 0.1320 & \textbf{0.1448} & 5.0\%  \\
                                                    & METEOR  & 0.1614       & 0.1655 & {\ul 0.1663} & 0.1574       & 0.1446 & \textbf{0.1804} & 11.8\% \\
\midrule
\multirow{2}{*}{LaMP:   News Headline Generation}   & ROUGE-L & 0.0398       & 0.0403 & 0.0403       & {\ul 0.0407} & 0.0401 & \textbf{0.0411} & 3.2\%  \\
                                                    & METEOR  & 0.0790       & 0.0792 & 0.0807       & {\ul 0.0800} & 0.7910 & \textbf{0.0809} & 2.5\%  \\
\midrule
\multirow{2}{*}{LaMP:   Scholarly Title Generation} & ROUGE-L & 0.1086       & 0.0909 & 0.0919       & {\ul 0.1100} & 0.1047 & \textbf{0.1366} & 25.8\% \\
                                                    & METEOR  & 0.2337       & 0.2066 & 0.2086       & {\ul 0.2348} & 0.1930 & \textbf{0.2575} & 10.2\% \\
\midrule
\multirow{2}{*}{LaMP:   Tweet Paraphrasing}         & ROUGE-L & 0.2506       & 0.2554 & {\ul 0.2571} & 0.2341       & 0.2204 & \textbf{0.2827} & 12.8\% \\
                                                    & METEOR  & 0.2588       & 0.2603 & {\ul 0.2634} & 0.2503       & 0.2389 & \textbf{0.3042} & 17.5\%
\\
                      \bottomrule
\end{tabular}
\end{adjustbox}
\caption{The performance results on LongLaMP and LaMP personalized text generation benchmarks. The best score is in \textbf{bold} and the second best is \underline{underlined}. }
\label{tab:main}
\end{table*}

\section{Experiments}
\subsection{Experimental Setup}
\paragraph{Benchmarks and Evaluation} 
We adopt LaMP benchmark~\cite{salemi-etal-2024-lamp} and LongLaMP benchmark~\cite{kumar2024longlamp}, which are designed for evaluating short-form and long-form personalized text generation, respectively. We exclude email generation tasks for both datasets since it involves private data that we cannot access. We choose the user split for both benchmarks and the dataset statistics are presented in Table~\ref{tab:datasets}. Following previous works~\cite{tan2024personalized, salemi2024comparing}, we use ROUGE-L and METEOR as evaluation metrics.

\paragraph{Baselines}
We compare our proposed \themodel with RAG-based personalization methods and PEFT-based personalization methods.

For RAG-based personalization, we employ two widely-used retrievers BM25~\cite{robertson2009probabilistic} and Contriever~\cite{lei2023unsupervised}. 

For PEFT-based personalization, we fine-tune user-specific LoRA adapter~\cite{hu2021lora} for each user using their profile $P_u=(x_i, y_i)_{i=0}^{|P_u|}$, using SFT loss in Equation~\ref{eq:sft}. Additionally, since we obtained style-agnostic responses, we also employ DPO loss~\cite{rafailov2024direct} to guide the model to generate user-authentic responses rather than style-agnostic responses.

\paragraph{Implementation Details}
We implement our proposed \themodel and all baselines with Llama-2-7B-chat~\cite{touvron2023llama}. For the RAG approach, we set the number of retrieved documents \( k = 2 \); for the PEFT approach, we set the rank of LoRA to 8. For \themodel, unless otherwise specified, we will use gpt-3.5-turbo to generate style-neutral responses and employ the simplest mean difference extracting function. We demonstrate the performance of using different extracting functions and general models in Appendix. We conduct experiments on the validation set to select the appropriate number of intervention layer $\ell$ and intervention strength $\alpha$ for each task. For more details, please refer to Appendix~\ref{app:exp_details}.

\begin{table}[h]
\centering
\begin{adjustbox}{max width=0.95\linewidth}
\begin{tabular}{@{}ccccc@{}}
\toprule
Task                                 & Averaged Cost $\downarrow$                & SFT & RAG  & Ours  \\ \midrule
\multirow{4}{*}{\makecell{Abstract \\ Generation}} & TT/User (s)    & 131.98   & \textbf{0.64}  & 27.23          \\
                                     & IL/Query (s)   & 22.59    & 18.90          & \textbf{15.59} \\
                                     & IL/5-Query (s) & 94.75    & 96.97          & \textbf{79.23} \\
                                     & SS/User (MB)   & 17.00    & 0.35           & \textbf{0.01}  \\ \midrule
\multirow{4}{*}{\makecell{Review \\ Generation}}   & TT/User (s)    & 62.45    & \textbf{0.44}  & 11.65          \\
                                     & IL/Query (s)   & 18.88    & \textbf{8.23}  & 11.75          \\
                                     & IL/5-Query (s) & 77.33    & \textbf{52.52} & 59.69          \\
                                     & SS/User (MB)   & 17.00    & 0.10           & \textbf{0.01}  \\ \midrule
\multirow{4}{*}{\makecell{News Headline \\ Generation}}            & TT/User (s)    & 123.28   & \textbf{1.22}  & 22.16          \\
                                     & IL/Query (s)   & 25.52    & 12.47          & \textbf{10.32} \\
                                     & IL/5-Query (s) & 105.00   & 78.08          & \textbf{57.80} \\
                                     & SS/User (MB)   & 17.00    & 0.83           & \textbf{0.01}  \\ \midrule
\multirow{4}{*}{\makecell{Scholarly Title \\ Generation}}               & TT/User (s)    & 112.31   & \textbf{0.51}  & 22.53          \\
                                     & IL/Query (s)   & 25.43    & \textbf{9.52}  & 10.49          \\
                                     & IL/5-Query (s) & 104.33   & \textbf{50.68} & 54.30          \\
                                     & SS/User (MB)   & 17.00    & 0.26           & \textbf{0.01}  \\ \bottomrule
\end{tabular}
\end{adjustbox}
\caption{Comparison of Training Time (TT, for train-free RAG and \themodel, represents pre-processing time), Inference Latency (IL) and Storage Space (SS) requirements across different methods. The lowest cost in in \textbf{bold}.}
\label{tab:exp-cost}
\end{table}

\subsection{Main Results}
\label{sec:cost}
\label{sec:layer}
By comparing our method with the baseline in terms of generation performance and efficiency, we demonstrate that our approach can achieve strong generation performance while maintaining high efficiency.
\paragraph{Generation Performance Comparison}
Table~\ref{tab:main} shows the generation performance comparison and  We can observe that:
\begin{itemize}[noitemsep, topsep=0pt]
    \item {\themodel demonstrates superior performance across both short-term and long-term personalized text generation tasks}. Notably,  \themodel achieves averaged 11\% and 8\% relative improvements on ROUGE-L and METEOR compared with RAG-based methods and PEFT-based methods, respectively.
    \item Both RAG-based and PEFT-based methods show unstable performance and cannot consistently improve base model across all tasks. RAG-based methods are more effective in tasks with less user history (review generation and tweet paraphrasing are the two tasks with the least user history), while PEFT performs better in scenarios with more historical data as it provides more training texts.
\end{itemize}

\paragraph{Efficiency Comparison}
Table~\ref{tab:exp-cost} shows the scalability comparison, where we implement Contriver as the retriever of RAG.
\begin{itemize}[noitemsep, topsep=0pt]
    \item In terms of training time, our method is training-free and requires only 1/5 of the preprocessing time compared to SFT. However, since RAG uses smaller retrievers (e.g., the Contriever model we use is no larger than 0.1B), RAG's preprocessing time is the shortest.
    \item In terms of inference latency, RAG is faster on tasks with less user history, but it becomes significantly slower on tasks with more user history. SFT takes too long to load and merge LoRA, making it unsuitable for scenarios that require frequent updates. Our method is independent of user history and does not require prolonged loading, making it a more versatile approach.
    \item In terms of storage space, our method only requires storing a single vector per user, making it unquestionably the most space-efficient, which occupies about $1 / 1700$ of the space required by SFT.
\end{itemize} 

\begin{figure}[t]
    \centering
    \includegraphics[width=\linewidth]{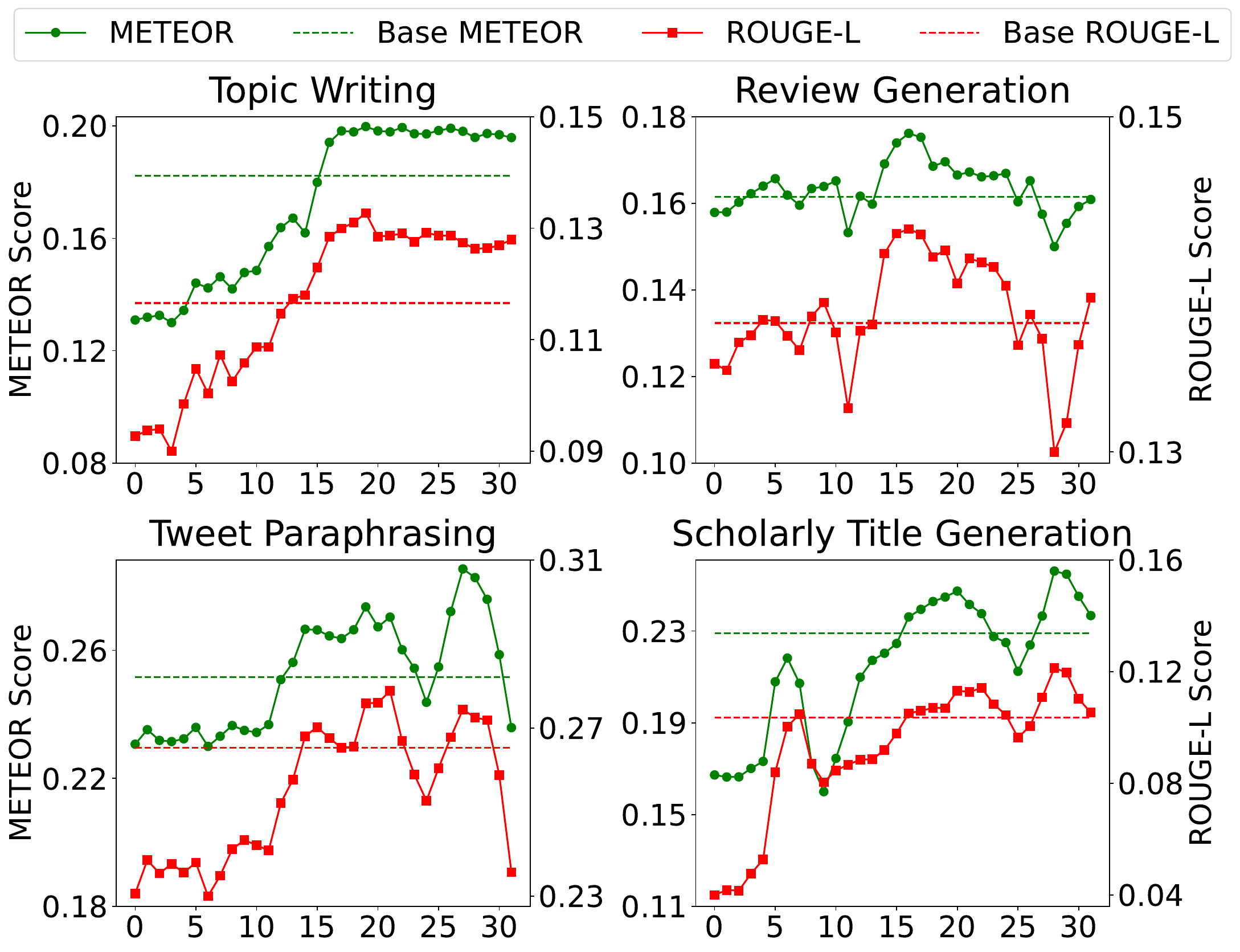} 
    \caption{Performance comparison across different intervention layers $l$.}
    \label{fig:layer}
\end{figure}
\subsection{Steering Analysis}
\label{sec:exp_layer}
\paragraph{Analysis of layers and multipliers}
The intervention layer $\ell$ and the intervention strength $\alpha$ are two important hyperparameters of our method. In this section, we analyze the impact of different values of $\ell$ and $\alpha$ on generation performance. The results are  shown in Figure~\ref{fig:layer} and Figure~\ref{fig:multiplier}, from which we can observe that:
\begin{itemize} [noitemsep, topsep=0pt]
    \item \textbf{The activations controlling the model's writing style are typically reflected in the middle to later layers}. As shown in Figure~\ref{fig:layer}, although there may be subtle differences across tasks, in general, the most effective intervention occurs when modifying the middle to later layers of the model (around layer 15 and beyond). Linear probing results in Section~\ref{sec:probing} also lead to the similar conclusion.
    \item \textbf{Positive intervention can guide the model to generate in the user's style, while negative intervention can push it away from that style.} As shown in Figure~\ref{fig:multiplier},  when \( \alpha < 0 \), the negative intervention causes the model's generated content to drift away from the user's style, resulting in a score lower than that of the non-personalized model. However, if \( \alpha \) is too large, it can cause abnormal activation values, thereby disrupting the generation process.
\end{itemize}

\begin{figure}[t]
    \centering
    \includegraphics[width=\linewidth]{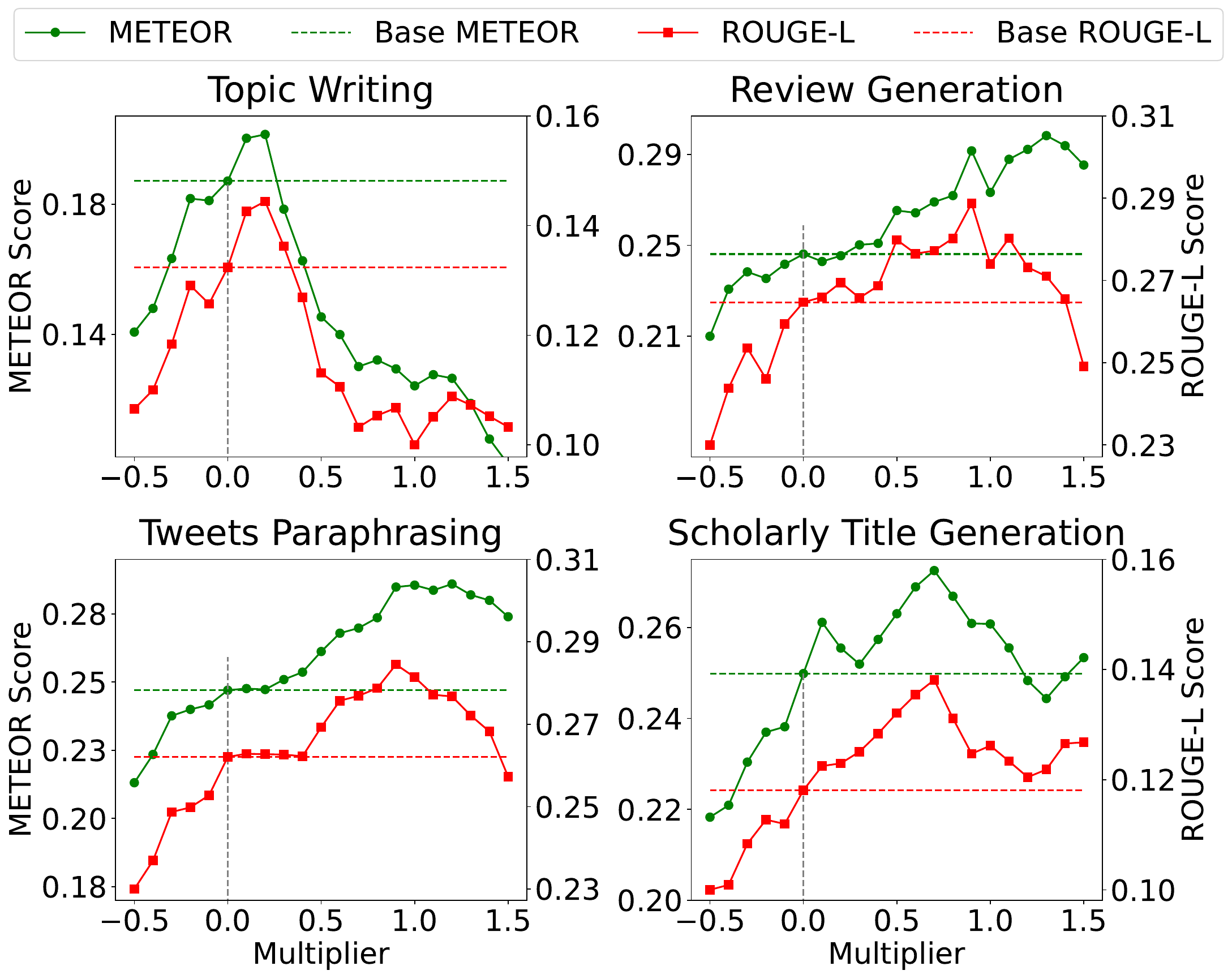} 
    \caption{Performance comparison across different intervention strengths $\alpha$.}
    \label{fig:multiplier}
\end{figure}

\subsection{Style Vector Analysis}
\label{sec:probing}
\paragraph{Probing Study}
To investigate how writing style features are encoded in the model's hidden states, we conduct a linear probing analysis across different layers of the base LLM. For each user $u \in \mathcal{U}$, we construct a binary classification task where the positive samples are the user's authentic historical texts $y_i \in \mathcal{P}_u$, and the negative samples are our framework's style-agnostic responses $\hat{y}_i$ generated for the same input contexts. We extract hidden states at layer $l$ for all samples and train a logistic regression classifier to distinguish between authentic and generated texts. Figure~\ref{fig:probing} shows the averaged probing results across all users, which reveals two key findings:
\begin{itemize}[noitemsep, topsep=0pt]
    \item \textbf{High Layer-wise Separability.} All layers achieve strong classification performance (AUC > 0.85), suggesting that user-specific stylistic patterns are robustly encoded throughout the network. This confirms our hypothesis that style information persists in the model's internal representations, even when not explicitly supervised.

    \item \textbf{The activations controlling the model's writing style are typically reflected in the middle to later layers}. The AUC increases with the depth of the layers, which aligns with our empirical findings in Section~\ref{sec:exp_layer}, where style steering interventions in these layers yielded optimal generation quality. The progressive feature refinement suggests that stylistic attributes are gradually distilled through the forward pass, reaching maximal linear separability in higher layers.
\end{itemize}

\begin{figure}[t]
    \centering
    \includegraphics[width=\linewidth]{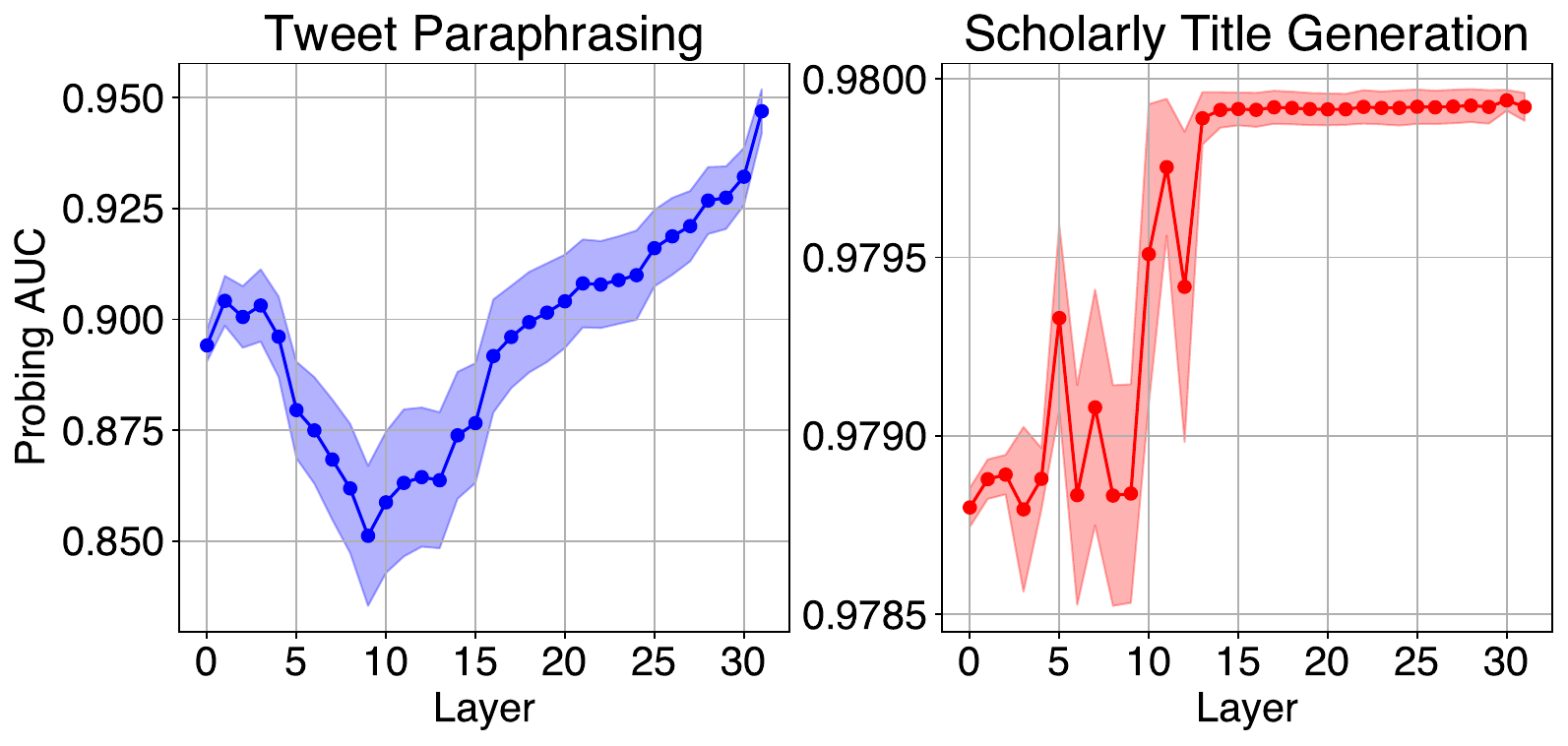} 
    \caption{Probing results on LaMP benchmark.}
    \label{fig:probing}
\end{figure}

\begin{figure*}[t]
    \centering
    \includegraphics[width=\linewidth]{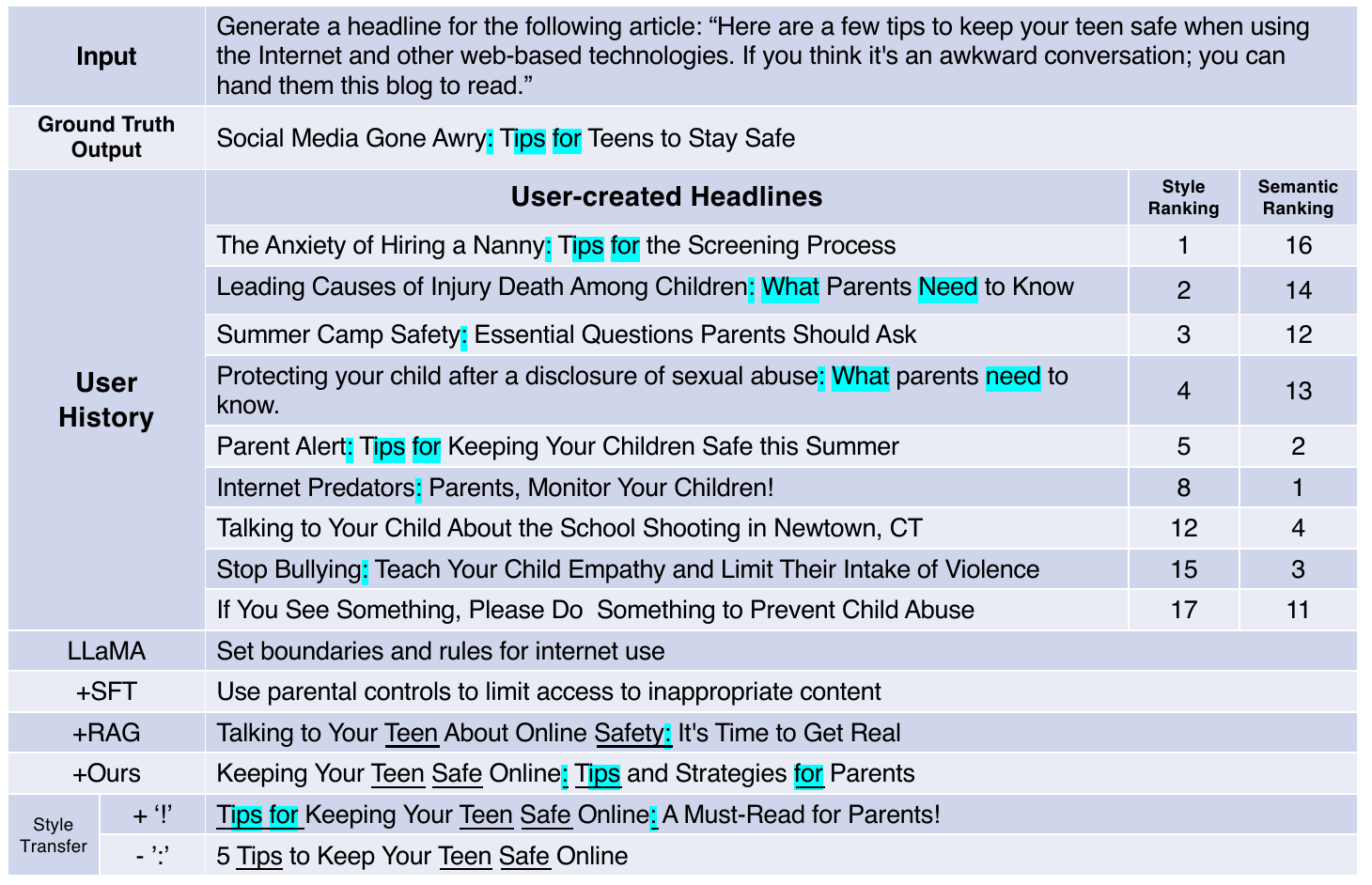} 
    \caption{Case study of user\_310 in News Headline Generation task. The \blue{highlighted} tokens are the top 5 tokens that most closely match the style vector among all historical tokens. The \underline{underline} words are the words that match the ground truth. 'Style Ranking' represents the ranking results based on the similarity between the historical headline embeddings with the style vector. 'Semantic Ranking' represents the ranking results obtained by Contriver~\cite{lei2023unsupervised}.}
    \label{fig:case}
\end{figure*}

\subsection{Case Study}
\label{sec:discrete}

To demonstrate the effectiveness of our method, we analyze a representative case from user\_310 in LaMP: News Headline Generation benchmark in Figure~\ref{fig:case}, demonstrating three key insights about our style vector approach:
\begin{itemize}[noitemsep, topsep=0pt]
    \item \textbf{Style vector encodes user preferences}. The highlighted tokens are the top 5 tokens that most closely match the style vector among all historical tokens. We can observe that the top-5 tokens (":", "ips", "for", "What", "Need") in historical headlines reveal consistent stylistic patterns of using subtitles and combinations such as "tips for" or "what  need". 
    \item \textbf{Style vector can steer personalized generation}. Our method generates "Keeping Your Teen Safe Online: Tips and Strategies for Parents", which naturally incorporates 3 key style tokens (":", "ips", "for") while maintaining content fidelity. However, the generation by baselines can not match user style preferences.
    \item \textbf{It's necessary to decouple style from semantic}. We list style ranking and semantic ranking of each historical headline, where style ranking represents the ranking results based on the similarity between the historical headline embeddings with the style vector, and semantic ranking represents the ranking results obtained by Contriver~\cite{lei2023unsupervised}. We can observe that headlines with higher style rankings exhibit stronger alignment with user-preferred stylistic patterns. However, there exists significant divergence between style ranking and semantic ranking. For RAG-based methods, the semantic-dominated retrieved headlines fail to provide useful patterns about stylistic preferences.
    \item \textbf{Style Transfer}.We tried rewriting the user's historical texts in a certain style (by instructing GPT) to recalculate the style vector, in order to observe whether we can steer the model to generate in the desired style. We targeted two styles: "exclamatory tone, ending with an exclamation mark" and "removal of colons and subheadings." The results show that our method can achieve style transfer while maintaining semantic fidelity, further demonstrating that the style vector can indeed encode the user's writing style.
    \end{itemize}

\section{Related Work}
\subsection{Personalized Text Generation}
The rapid evolution of LLMs has fundamentally transformed content generation paradigms, shifting from generic outputs to sophisticated personalized text generation. 
Current methodologies in personalized generation predominantly fall into two technical categories: {Retrieval-Augmented Generation (RAG)} approaches leverage users' historical content ($P_u$) through dynamic retrieval mechanisms. While foundational work~\cite{zhang2023memory,salemi2024comparing,Salemi2024OptimizationMF,Richardson2023IntegratingSA} established basic retrieval frameworks, recent innovations have enhanced these paradigms. \citet{Richardson2023IntegratingSA,zhang-2024-guided,Tan2025CanLL} developed profile-augmented prompting strategies, while \citet{salemi2024learning} introduced feedback-driven retrieval model optimization, demonstrating improved personalization accuracy.
{Parameter-Efficient Fine-Tuning (PEFT)} methods adopt an alternative paradigm by adapting per-user parameters through lightweight adapter modules. Comparative studies~\cite{salemi2024comparing} reveal that PEFT approaches, particularly those employing user-specific adapter tuning~\cite{tan2024personalized,zhuang2024hydra,Liu2024LLMsP,Ding2025PersonalizedLM}, achieve competitive personalization while maintaining computational efficiency.

\subsection{Activation Engineering} 
Emerging research in activation engineering has uncovered that LLMs encode semantic concepts as linear subspaces within hidden activation representations~\cite{zou2023representation, liu2023context, rimsky-etal-2024-steering}. This geometric interpretation enables targeted behavioral steering through linear interventions during inference. \citet{turner2023activation} pioneered activation addition using contrastive-derived steering vectors for sentiment and topic control, while \citet{rimsky-etal-2024-steering} enhanced steering precision through mass-mean activation differentials. \citet{zhang2024truthx} identified truth-correlated heads via linear probing, achieving enhanced veracity through targeted modulation. Complementing this, \citet{chen2024truth} developed multi-directional orthogonal steering to amplify truthfulness in model responses.

\section{Conclusion}  
In this work, we demonstrate that user's writing style can be represented as a vector in LLM's activation-space. Based on this insight, we introduces a simple yet effective frame, \themodel, that achieves personalized text generation through inference time intervention, without parameter updates or retrievals. Experiments on both short- and long-form personalization benchmarks show our method can achieve strong generation performance while maintaining high efficiency.

\section*{Limitations}
\label{sec:limitations}

While our framework demonstrates significant advantages in efficiency and effectiveness, several limitations warrant discussion to guide future research:

Our training-free style vector derivation, though efficient, may not achieve optimal disentanglement of style from content. The current contrastive approach relies on the model's inherent ability to separate these features through simple activation arithmetic. Future work could explore hybrid approaches that combine our parametric-free method with lightweight optimization techniques to refine the style vectors while maintaining storage efficiency.

The single-vector user representation, while storage-efficient, potentially conflates multiple stylistic dimensions (e.g., lexical preferences, syntactic structures, and discourse patterns). A more granular approach could represent users through sparse combinations~\cite{cunningham2023sparse,lieberum2024gemma} of concept-specific vectors, enabling precise control over individual style components.  

 Our evaluation focuses on established benchmarks (LaMP and LongLaMP) that assume domain homogeneity within each user's historical data. However, real-world personalization scenarios often involve \textit{cross-domain style consistency} – users may employ distinct stylistic registers across different tasks (e.g., formal emails vs. casual social media posts). Current benchmarks lack the capability to assess whether learned style vectors can: (1) preserve task-appropriate stylistic variations within users, or (2) prevent negative interference between conflicting domain-specific patterns. Future work should develop cross-domain personalization benchmarks that incorporate mixed-task histories.

\section*{Ethics Statement}
\label{sec:ethics}
The experimental datasets are publicly available from some previous works, downloaded via official APIs. The information regarding users in all datasets has been anonymized, ensuring there are no privacy concerns related to the users. We do not disclose any non-open-source data, and we ensure that our actions comply with ethical standards. We use publicly available pre-trained models, i.e., LLaMA-2, Contriver, and APIs, i.e., GPT-3.5-turbo,  DeepSeek-Chat. All the checkpoints and datasets are carefully processed by their authors to ensure that there are no ethical problems.

\bibliography{acl}

\begin{thebibliography}{41}
\expandafter\ifx\csname natexlab\endcsname\relax\def\natexlab#1{#1}\fi

\bibitem[{Alhafni et~al.(2024)Alhafni, Kulkarni, Kumar, and Raheja}]{Alhafni2024PersonalizedTG}
Bashar Alhafni, Vivek Kulkarni, Dhruv Kumar, and Vipul Raheja. 2024.
\newblock Personalized text generation with fine-grained linguistic control.
\newblock \emph{ArXiv}.

\bibitem[{Au et~al.(2025)Au, Dimacali, Pedirappagari, Park, Dernoncourt, Wang, Kanakaris, Deilamsalehy, Rossi, and Ahmed}]{au2025personalized}
Steven Au, Cameron~J Dimacali, Ojasmitha Pedirappagari, Namyong Park, Franck Dernoncourt, Yu~Wang, Nikos Kanakaris, Hanieh Deilamsalehy, Ryan~A Rossi, and Nesreen~K Ahmed. 2025.
\newblock Personalized graph-based retrieval for large language models.
\newblock \emph{arXiv preprint arXiv:2501.02157}.

\bibitem[{Cai et~al.(2024)Cai, Li, Wang, Zhu, Shen, Li, and Chua}]{cai2024large}
Hongru Cai, Yongqi Li, Wenjie Wang, Fengbin Zhu, Xiaoyu Shen, Wenjie Li, and Tat-Seng Chua. 2024.
\newblock Large language models empowered personalized web agents.
\newblock \emph{arXiv preprint arXiv:2410.17236}.

\bibitem[{Chen et~al.(2024{\natexlab{a}})Chen, Liu, Huang, Wu, Liu, Jiang, Pu, Lei, Chen, Wang et~al.}]{chen2024large}
Jin Chen, Zheng Liu, Xu~Huang, Chenwang Wu, Qi~Liu, Gangwei Jiang, Yuanhao Pu, Yuxuan Lei, Xiaolong Chen, Xingmei Wang, et~al. 2024{\natexlab{a}}.
\newblock When large language models meet personalization: Perspectives of challenges and opportunities.
\newblock \emph{World Wide Web}, 27(4):42.

\bibitem[{Chen et~al.(2024{\natexlab{b}})Chen, Sun, Jiao, Lian, Kang, Wang, and Xu}]{chen2024truth}
Zhongzhi Chen, Xingwu Sun, Xianfeng Jiao, Fengzong Lian, Zhanhui Kang, Di~Wang, and Chengzhong Xu. 2024{\natexlab{b}}.
\newblock Truth forest: Toward multi-scale truthfulness in large language models through intervention without tuning.
\newblock In \emph{Proceedings of the AAAI Conference on Artificial Intelligence}, volume~38, pages 20967--20974.

\bibitem[{Chen and Moscholios(2024)}]{Chen2024UsingPT}
Ziyang Chen and Stylios Moscholios. 2024.
\newblock Using prompts to guide large language models in imitating a real person's language style.
\newblock \emph{ArXiv}.

\bibitem[{Cunningham et~al.(2023)Cunningham, Ewart, Riggs, Huben, and Sharkey}]{cunningham2023sparse}
Hoagy Cunningham, Aidan Ewart, Logan Riggs, Robert Huben, and Lee Sharkey. 2023.
\newblock Sparse autoencoders find highly interpretable features in language models.
\newblock \emph{arXiv preprint arXiv:2309.08600}.

\bibitem[{Ding et~al.(2025)Ding, Tan, Liu, Niu, Meng, Zhou, Liu, Wu, and Chen}]{Ding2025PersonalizedLM}
Yucheng Ding, Yangwenjian Tan, Xiangyu Liu, Chaoyue Niu, Fandong Meng, Jie Zhou, Ning Liu, Fan Wu, and Guihai Chen. 2025.
\newblock Personalized language model learning on text data without user identifiers.
\newblock \emph{ArXiv}.

\bibitem[{Fisher et~al.(2024)Fisher, Hallinan, Lu, Gordon, Harchaoui, and Choi}]{Fisher2024StyleRemixIA}
Jillian~R. Fisher, Skyler Hallinan, Ximing Lu, Mitchell Gordon, Zaid Harchaoui, and Yejin Choi. 2024.
\newblock \href {https://api.semanticscholar.org/CorpusID:271974917} {Styleremix: Interpretable authorship obfuscation via distillation and perturbation of style elements}.
\newblock \emph{ArXiv}, abs/2408.15666.

\bibitem[{Hu et~al.(2021)Hu, Shen, Wallis, Allen-Zhu, Li, Wang, Wang, and Chen}]{hu2021lora}
Edward~J Hu, Yelong Shen, Phillip Wallis, Zeyuan Allen-Zhu, Yuanzhi Li, Shean Wang, Lu~Wang, and Weizhu Chen. 2021.
\newblock Lora: Low-rank adaptation of large language models.
\newblock \emph{arXiv preprint arXiv:2106.09685}.

\bibitem[{Jang et~al.(2023)Jang, Kim, Lin, Wang, Hessel, Zettlemoyer, Hajishirzi, Choi, and Ammanabrolu}]{jang2023personalized}
Joel Jang, Seungone Kim, Bill~Yuchen Lin, Yizhong Wang, Jack Hessel, Luke Zettlemoyer, Hannaneh Hajishirzi, Yejin Choi, and Prithviraj Ammanabrolu. 2023.
\newblock Personalized soups: Personalized large language model alignment via post-hoc parameter merging.
\newblock \emph{arXiv preprint arXiv:2310.11564}.

\bibitem[{Kirk et~al.(2024)Kirk, Vidgen, R{\"o}ttger, and Hale}]{kirk2024benefits}
Hannah~Rose Kirk, Bertie Vidgen, Paul R{\"o}ttger, and Scott~A Hale. 2024.
\newblock The benefits, risks and bounds of personalizing the alignment of large language models to individuals.
\newblock \emph{Nature Machine Intelligence}, pages 1--10.

\bibitem[{Kumar et~al.(2024)Kumar, Viswanathan, Yerra, Salemi, Rossi, Dernoncourt, Deilamsalehy, Chen, Zhang, Agarwal et~al.}]{kumar2024longlamp}
Ishita Kumar, Snigdha Viswanathan, Sushrita Yerra, Alireza Salemi, Ryan~A Rossi, Franck Dernoncourt, Hanieh Deilamsalehy, Xiang Chen, Ruiyi Zhang, Shubham Agarwal, et~al. 2024.
\newblock Longlamp: A benchmark for personalized long-form text generation.
\newblock \emph{arXiv preprint arXiv:2407.11016}.

\bibitem[{Lei et~al.(2023)Lei, Ding, Cao, Zan, Yates, and Tao}]{lei2023unsupervised}
Yibin Lei, Liang Ding, Yu~Cao, Changtong Zan, Andrew Yates, and Dacheng Tao. 2023.
\newblock Unsupervised dense retrieval with relevance-aware contrastive pre-training.
\newblock In \emph{Findings of the Association for Computational Linguistics: ACL 2023}, pages 10932--10940.

\bibitem[{Lieberum et~al.(2024)Lieberum, Rajamanoharan, Conmy, Smith, Sonnerat, Varma, Kram{\'a}r, Dragan, Shah, and Nanda}]{lieberum2024gemma}
Tom Lieberum, Senthooran Rajamanoharan, Arthur Conmy, Lewis Smith, Nicolas Sonnerat, Vikrant Varma, J{\'a}nos Kram{\'a}r, Anca Dragan, Rohin Shah, and Neel Nanda. 2024.
\newblock Gemma scope: Open sparse autoencoders everywhere all at once on gemma 2.
\newblock \emph{arXiv preprint arXiv:2408.05147}.

\bibitem[{Lin et~al.(2024)Lin, Wang, Pan, Manjunatha, Rossi, Lau, Huang, and Sun}]{lin2024persona}
Zihao Lin, Zichao Wang, Yuanting Pan, Varun Manjunatha, Ryan Rossi, Angela Lau, Lifu Huang, and Tong Sun. 2024.
\newblock Persona-sq: A personalized suggested question generation framework for real-world documents.
\newblock \emph{arXiv preprint arXiv:2412.12445}.

\bibitem[{Liu et~al.(2024{\natexlab{a}})Liu, Gu, Zheng, Xiang, Wu, Fu, and He}]{Liu2024DynamicGO}
Jianzhi Liu, Hexiang Gu, Tianyu Zheng, Liuyu Xiang, Huijia Wu, Jie Fu, and Zhaofeng He. 2024{\natexlab{a}}.
\newblock Dynamic generation of personalities with large language models.
\newblock \emph{ArXiv}, abs/2404.07084.

\bibitem[{Liu et~al.(2024{\natexlab{b}})Liu, Zhu, Wang, Wei, Min, Lu, Wang, Yin, and Dou}]{Liu2024LLMsP}
Jiongnan Liu, Yutao Zhu, Shuting Wang, Xiaochi Wei, Erxue Min, Yu~Lu, Shuaiqiang Wang, Dawei Yin, and Zhicheng Dou. 2024{\natexlab{b}}.
\newblock Llms + persona-plug = personalized llms.
\newblock \emph{ArXiv}.

\bibitem[{Liu et~al.(2023)Liu, Ye, Xing, and Zou}]{liu2023context}
Sheng Liu, Haotian Ye, Lei Xing, and James Zou. 2023.
\newblock In-context vectors: Making in context learning more effective and controllable through latent space steering.
\newblock \emph{arXiv preprint arXiv:2311.06668}.

\bibitem[{Paszke et~al.(2019)Paszke, Gross, Massa, Lerer, Bradbury, Chanan, Killeen, Lin, Gimelshein, Antiga et~al.}]{paszke2019pytorch}
Adam Paszke, Sam Gross, Francisco Massa, Adam Lerer, James Bradbury, Gregory Chanan, Trevor Killeen, Zeming Lin, Natalia Gimelshein, Luca Antiga, et~al. 2019.
\newblock Pytorch: An imperative style, high-performance deep learning library.
\newblock \emph{Advances in neural information processing systems}, 32.

\bibitem[{Rafailov et~al.(2024)Rafailov, Sharma, Mitchell, Manning, Ermon, and Finn}]{rafailov2024direct}
Rafael Rafailov, Archit Sharma, Eric Mitchell, Christopher~D Manning, Stefano Ermon, and Chelsea Finn. 2024.
\newblock Direct preference optimization: Your language model is secretly a reward model.
\newblock \emph{Advances in Neural Information Processing Systems}, 36.

\bibitem[{Richardson et~al.(2023)Richardson, Zhang, Gillespie, Kar, Singh, Raeesy, Khan, and Sethy}]{Richardson2023IntegratingSA}
Chris Richardson, Yao Zhang, Kellen Gillespie, Sudipta Kar, Arshdeep Singh, Zeynab Raeesy, Omar~Zia Khan, and Abhinav Sethy. 2023.
\newblock Integrating summarization and retrieval for enhanced personalization via large language models.
\newblock \emph{ArXiv}.

\bibitem[{Rimsky et~al.(2024)Rimsky, Gabrieli, Schulz, Tong, Hubinger, and Turner}]{rimsky-etal-2024-steering}
Nina Rimsky, Nick Gabrieli, Julian Schulz, Meg Tong, Evan Hubinger, and Alexander Turner. 2024.
\newblock \href {https://doi.org/10.18653/v1/2024.acl-long.828} {Steering llama 2 via contrastive activation addition}.
\newblock In \emph{Proceedings of the 62nd Annual Meeting of the Association for Computational Linguistics (Volume 1: Long Papers)}, pages 15504--15522, Bangkok, Thailand. Association for Computational Linguistics.

\bibitem[{Robertson et~al.(2009)Robertson, Zaragoza et~al.}]{robertson2009probabilistic}
Stephen Robertson, Hugo Zaragoza, et~al. 2009.
\newblock The probabilistic relevance framework: Bm25 and beyond.
\newblock \emph{Foundations and Trends{\textregistered} in Information Retrieval}, 3(4):333--389.

\bibitem[{Salemi et~al.(2024{\natexlab{a}})Salemi, Kallumadi, and Zamani}]{Salemi2024OptimizationMF}
Alireza Salemi, Surya Kallumadi, and Hamed Zamani. 2024{\natexlab{a}}.
\newblock Optimization methods for personalizing large language models through retrieval augmentation.
\newblock In \emph{Annual International ACM SIGIR Conference on Research and Development in Information Retrieval}.

\bibitem[{Salemi et~al.(2024{\natexlab{b}})Salemi, Mysore, Bendersky, and Zamani}]{salemi-etal-2024-lamp}
Alireza Salemi, Sheshera Mysore, Michael Bendersky, and Hamed Zamani. 2024{\natexlab{b}}.
\newblock \href {https://doi.org/10.18653/v1/2024.acl-long.399} {{L}a{MP}: When large language models meet personalization}.
\newblock In \emph{Proceedings of the 62nd Annual Meeting of the Association for Computational Linguistics (Volume 1: Long Papers)}, pages 7370--7392, Bangkok, Thailand. Association for Computational Linguistics.

\bibitem[{Salemi and Zamani(2024{\natexlab{a}})}]{salemi2024comparing}
Alireza Salemi and Hamed Zamani. 2024{\natexlab{a}}.
\newblock Comparing retrieval-augmentation and parameter-efficient fine-tuning for privacy-preserving personalization of large language models.
\newblock \emph{arXiv preprint arXiv:2409.09510}.

\bibitem[{Salemi and Zamani(2024{\natexlab{b}})}]{salemi2024learning}
Alireza Salemi and Hamed Zamani. 2024{\natexlab{b}}.
\newblock Learning to rank for multiple retrieval-augmented models through iterative utility maximization.
\newblock \emph{arXiv preprint arXiv:2410.09942}.

\bibitem[{Tan et~al.(2024{\natexlab{a}})Tan, Liu, and Jiang}]{tan2024personalized}
Zhaoxuan Tan, Zheyuan Liu, and Meng Jiang. 2024{\natexlab{a}}.
\newblock Personalized pieces: Efficient personalized large language models through collaborative efforts.
\newblock \emph{arXiv preprint arXiv:2406.10471}.

\bibitem[{Tan et~al.(2024{\natexlab{b}})Tan, Zeng, Tian, Liu, Yin, and Jiang}]{tan-etal-2024-democratizing}
Zhaoxuan Tan, Qingkai Zeng, Yijun Tian, Zheyuan Liu, Bing Yin, and Meng Jiang. 2024{\natexlab{b}}.
\newblock Democratizing large language models via personalized parameter-efficient fine-tuning.
\newblock In \emph{Proceedings of the 2024 Conference on Empirical Methods in Natural Language Processing}.

\bibitem[{Tan et~al.(2025)Tan, Zeng, Zeng, Wu, Liu, Mo, and Jiang}]{Tan2025CanLL}
Zhaoxuan Tan, Zinan Zeng, Qingkai Zeng, Zhenyu Wu, Zheyuan Liu, Fengran Mo, and Meng Jiang. 2025.
\newblock Can large language models understand preferences in personalized recommendation?
\newblock \emph{ArXiv}.

\bibitem[{Touvron et~al.(2023)Touvron, Martin, Stone, Albert, Almahairi, Babaei, Bashlykov, Batra, Bhargava, Bhosale et~al.}]{touvron2023llama}
Hugo Touvron, Louis Martin, Kevin Stone, Peter Albert, Amjad Almahairi, Yasmine Babaei, Nikolay Bashlykov, Soumya Batra, Prajjwal Bhargava, Shruti Bhosale, et~al. 2023.
\newblock Llama 2: Open foundation and fine-tuned chat models.
\newblock \emph{arXiv preprint arXiv:2307.09288}.

\bibitem[{Turner et~al.(2023)Turner, Thiergart, Leech, Udell, Vazquez, Mini, and MacDiarmid}]{turner2023activation}
Alexander~Matt Turner, Lisa Thiergart, Gavin Leech, David Udell, Juan~J Vazquez, Ulisse Mini, and Monte MacDiarmid. 2023.
\newblock Activation addition: Steering language models without optimization.
\newblock \emph{arXiv e-prints}, pages arXiv--2308.

\bibitem[{Wolf(2019)}]{wolf2019huggingface}
T~Wolf. 2019.
\newblock Huggingface's transformers: State-of-the-art natural language processing.
\newblock \emph{arXiv preprint arXiv:1910.03771}.

\bibitem[{Zhang(2024)}]{zhang-2024-guided}
Jiarui Zhang. 2024.
\newblock Guided profile generation improves personalization with large language models.
\newblock In \emph{Findings of the Association for Computational Linguistics: EMNLP 2024}.

\bibitem[{Zhang et~al.(2023)Zhang, Zhao, Kang, and Liu}]{zhang2023memory}
Kai Zhang, Fubang Zhao, Yangyang Kang, and Xiaozhong Liu. 2023.
\newblock Memory-augmented llm personalization with short-and long-term memory coordination.
\newblock \emph{arXiv preprint arXiv:2309.11696}.

\bibitem[{Zhang et~al.(2024{\natexlab{a}})Zhang, Yu, and Feng}]{zhang2024truthx}
Shaolei Zhang, Tian Yu, and Yang Feng. 2024{\natexlab{a}}.
\newblock Truthx: Alleviating hallucinations by editing large language models in truthful space.
\newblock \emph{arXiv preprint arXiv:2402.17811}.

\bibitem[{Zhang et~al.(2024{\natexlab{b}})Zhang, Rossi, Kveton, Shao, Yang, Zamani, Dernoncourt, Barrow, Yu, Kim et~al.}]{zhang2024personalization}
Zhehao Zhang, Ryan~A Rossi, Branislav Kveton, Yijia Shao, Diyi Yang, Hamed Zamani, Franck Dernoncourt, Joe Barrow, Tong Yu, Sungchul Kim, et~al. 2024{\natexlab{b}}.
\newblock Personalization of large language models: A survey.
\newblock \emph{arXiv preprint arXiv:2411.00027}.

\bibitem[{Zhu et~al.(2025)Zhu, Jin, and Coifman}]{Zhu2025InvestigatingLL}
Jianfeng Zhu, Ruoming Jin, and Karin~G. Coifman. 2025.
\newblock Investigating large language models in inferring personality traits from user conversations.
\newblock \emph{ArXiv}.

\bibitem[{Zhuang et~al.(2024)Zhuang, Sun, Yu, Qiang, Wang, Zhang, and Dai}]{zhuang2024hydra}
Yuchen Zhuang, Haotian Sun, Yue Yu, Rushi Qiang, Qifan Wang, Chao Zhang, and Bo~Dai. 2024.
\newblock Hydra: Model factorization framework for black-box llm personalization.
\newblock \emph{arXiv preprint arXiv:2406.02888}.

\bibitem[{Zou et~al.(2023)Zou, Phan, Chen, Campbell, Guo, Ren, Pan, Yin, Mazeika, Dombrowski et~al.}]{zou2023representation}
Andy Zou, Long Phan, Sarah Chen, James Campbell, Phillip Guo, Richard Ren, Alexander Pan, Xuwang Yin, Mantas Mazeika, Ann-Kathrin Dombrowski, et~al. 2023.
\newblock Representation engineering: A top-down approach to ai transparency.
\newblock \emph{arXiv preprint arXiv:2310.01405}.

\end{thebibliography}

\newpage

\appendix
\section{Algorithm} 
The complete procedure is formalized in Algorithm~\ref{alg:style_steering}.
\begin{algorithm}[t]
\caption{Personalized Generation with Style Steering}
\label{alg:style_steering}
\begin{algorithmic}[1]
\Require 
    User interaction history \( P_u = \{(x_i, y_i)\}_{i=1}^{|P_u|} \),
    general LLM \( M_g \),
    intervention layer \( \ell \),
    scaling factor \( \alpha \),
    input query \( x \)
\Ensure 
    Personalized generation model \( M \)

\State \textbf{Stage 1: Generate Style-Agnostic Responses}
    \For{each data pair \((x_i, y_i) \in P_u\)}
        \State Generate style-agnostic response \(\hat{y}_i \gets M_g(x_i)\)
    \EndFor

\State \textbf{Stage 2: Extract Style Vector}
    \For{each data pair \((x_i, y_i) \in P_u\)}
        \State Compute positive activation \( a_{p,i}^\ell \gets h_\ell(x_i \oplus y_i) \)
        \State Compute negative activation \( a_{n,i}^\ell \gets h_\ell(x_i \oplus \hat{y}_i) \)
    \EndFor
    \State Extract style vector \( s_u^\ell \gets f(\{a_{p,i}^\ell, a_{n,i}^\ell\}_{i=1}^{|P_u|}) \)

\State \textbf{Stage 3: Activation Steering}
    \For{each generation position \( t \geq |x| \)} 
        \State Retrieve original activation \( h_\ell(x)_t \)
        \State Inject style vector \( h'_\ell(x)_t \gets h_\ell(x)_t + \alpha s_u^\ell \)
    \EndFor
\end{algorithmic}
\end{algorithm}

\section{Experiment Details}
\label{app:exp_details}

\paragraph{DPO Baseline}
The DPO algorithm \cite{rafailov2024direct} reframes preference learning by directly optimizing a policy to align with human preferences without explicit reward modeling. Since we obtained style-agnostic responses, we also employ DPO loss~\cite{rafailov2024direct} to guide the model to generate user-authentic responses $y_i$ rather than style-agnostic responses $\hat{y}_i$.
\begin{equation}
\label{eq:dpo}
\begin{split}
\theta_u^* =  \arg\min_{\theta} \sum_{(x_i,y_i,\hat{y}_i) \in P_u} 
& -\log \sigma\bigg( \beta \log \frac{M_\theta(y_i \mid x_i)}{M_{\text{ref}}(y_i \mid x_i)} \\
& - \beta \log \frac{M_\theta(\hat{y}_i \mid x_i)}{M_{\text{ref}}(\hat{y}_i \mid x_i)} \bigg)
\end{split}
\end{equation}
where $M_\theta$ is the policy with adapter $\theta_u$, $M_{\text{ref}}$ is the reference policy (base model $M$ with frozen parameters), $\sigma$ denotes the sigmoid function, and $\beta$ controls deviation from the reference policy. This approach enables parameter-efficient preference alignment through lightweight adapters while maintaining the base model's capabilities.

\paragraph{Implementation Details}
All experiments were performed on a cluster of 8 NVIDIA RTX 3090 GPUs, with implementations built upon the PyTorch framework~\cite{paszke2019pytorch}, HuggingFace Transformers~\cite{wolf2019huggingface} library. To save computational resources, we apply 8-bit quantization and greedy decoding for all methods.

\section{Additional Experimental Analysis}
\subsection{Analysis of Extracting Function and Intervention Position}
\label{app:funcs_positions}
We compare three different extracting functions in Equation~\ref{eq:agg} and different intervention token positions $t$ in Equation~\ref{eq:steer}. We use three different intervention positions: intervening on all input tokens, intervening only on the last input token, and intervening on each newly generated token. The results are shown in Figure~\ref{fig:pos_agg}. As we can see, using any extracting function and intervention position results in significant improvements in personalized text generation. Although it is very simple and does not introduce excessive complexity, the performance of the Mean Difference function is still highly superior. Moreover, the more tokens are intervened, the more pronounced the performance improvement.

\begin{figure}[t]
    \centering
    \includegraphics[width=\linewidth]{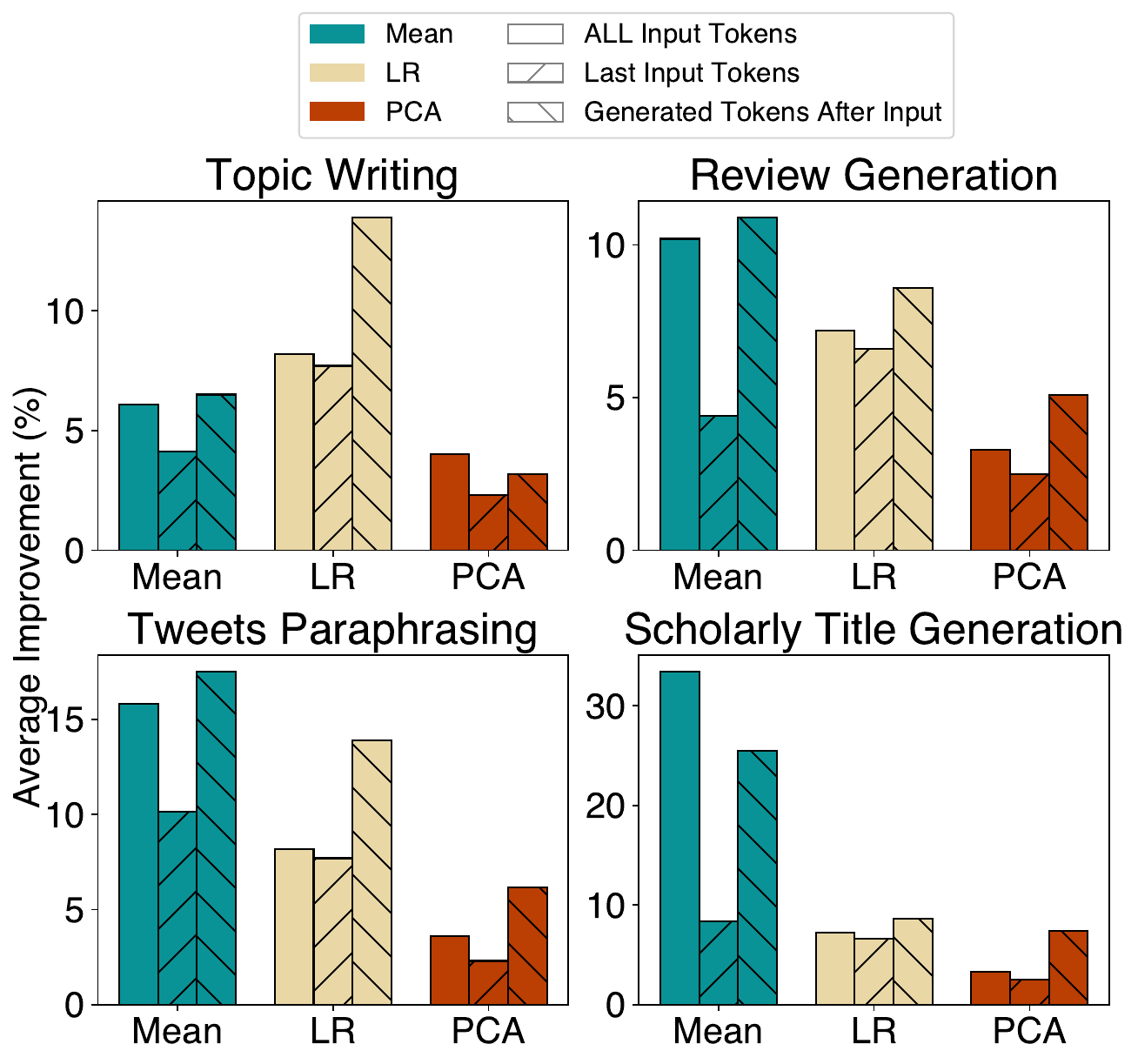} 
    \caption{Performance comparison across different extracting functions and intervention positions $t$.}
    \label{fig:pos_agg}
\end{figure}

\subsection{Analysis of General Model Selection}
We compare the different choices of the general LLM $M_g$ in Equation~\ref{eq:general_model} which is designed to generate style-agnostic responses. The results are shown in Figure~\ref{fig:base_model}, from which we can observe that the proposed \themodel is robust over different general models. The general model does not have to be the same as the model being intervened (LLaMA-2-7b); in fact, text generated by a more powerful model tends to have a higher relevance to the input $x$, greater diversity, and is more conducive to the extraction of style vectors.

\label{app:base_model}
\begin{figure}[t]
    \centering
    \includegraphics[width=\linewidth]{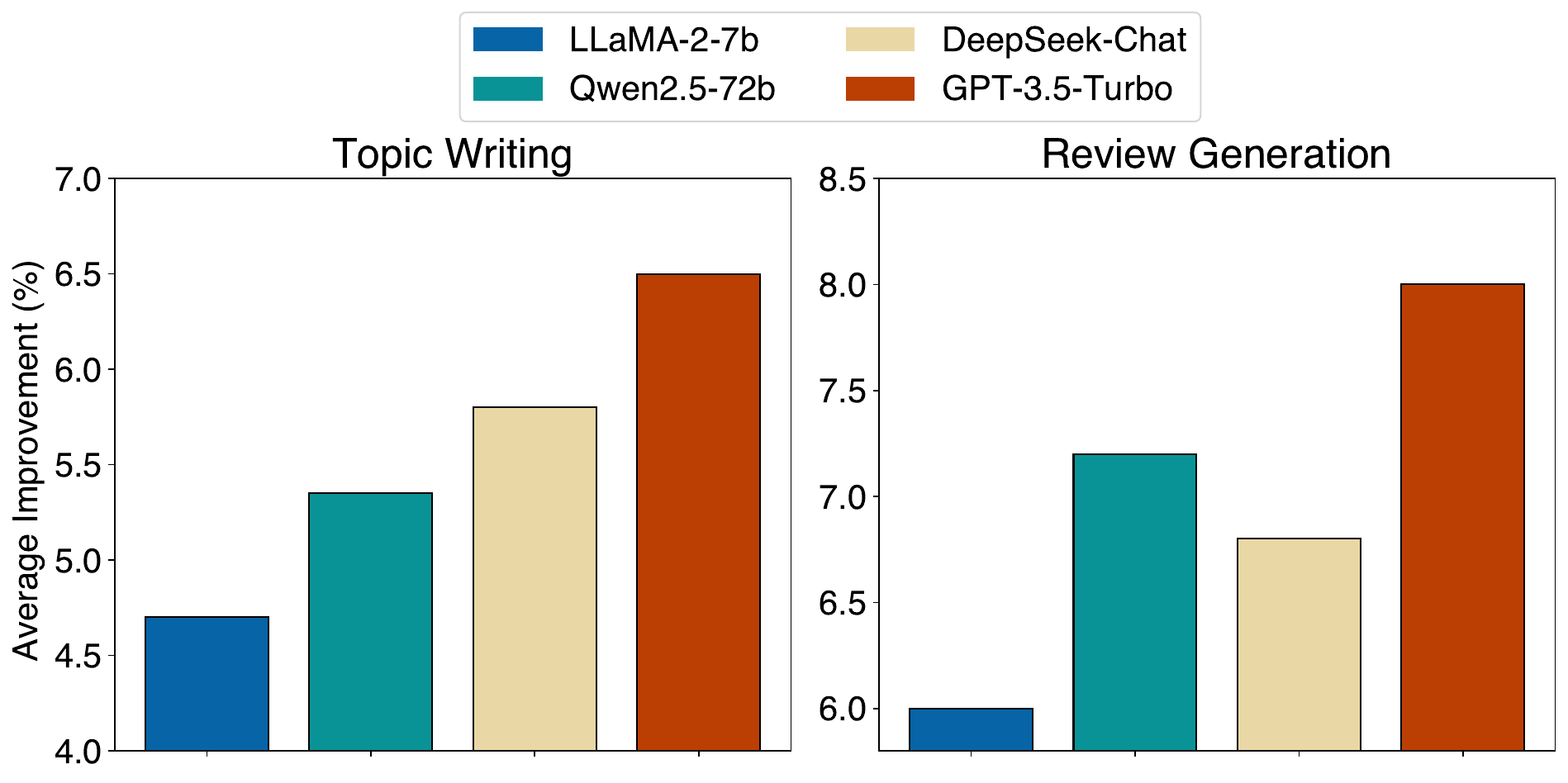} 
    \caption{Performance comparison with different generic models $M_g$.}
    \label{fig:base_model}
\end{figure}

\subsection{Clustering}
\label{app:cluster}
Figure~\ref{fig:cluster} illustrates the distribution of clustered style vectors for all users in two tasks of the LaMP benchmark. As can be seen, the dimensionality-reduced user style vectors can be grouped into several clusters, indicating that different users may share similar writing styles. 

Additionally, in Figure~\ref{fig:case_cluster}, we provide examples of some clusters and highlight the significant writing style patterns of these clusters. For example, in the case of cluster 1, the users within it share two writing style patterns: one prefers starting with numbers, and likes adding parentheses at the end to supplement the content. For cluster 2, all users share one pattern: they tend to use the dash '--' to connect elements in the title.

\begin{figure}[h]
    \centering
    \includegraphics[width=\linewidth]{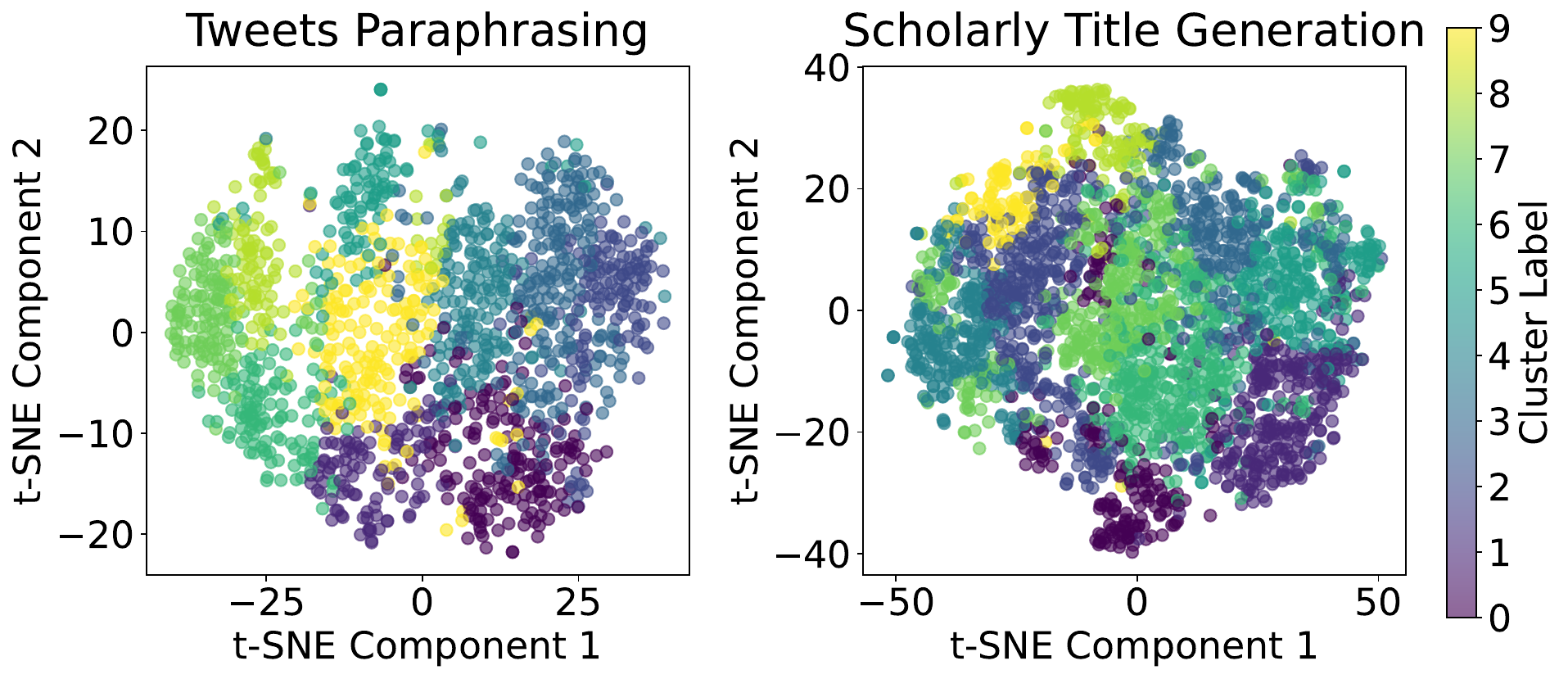} 
    \caption{Clustering results of style vectors of all users.}
    \label{fig:cluster}
\end{figure}

\begin{figure*}[t]
    \centering
    \includegraphics[width=\linewidth]{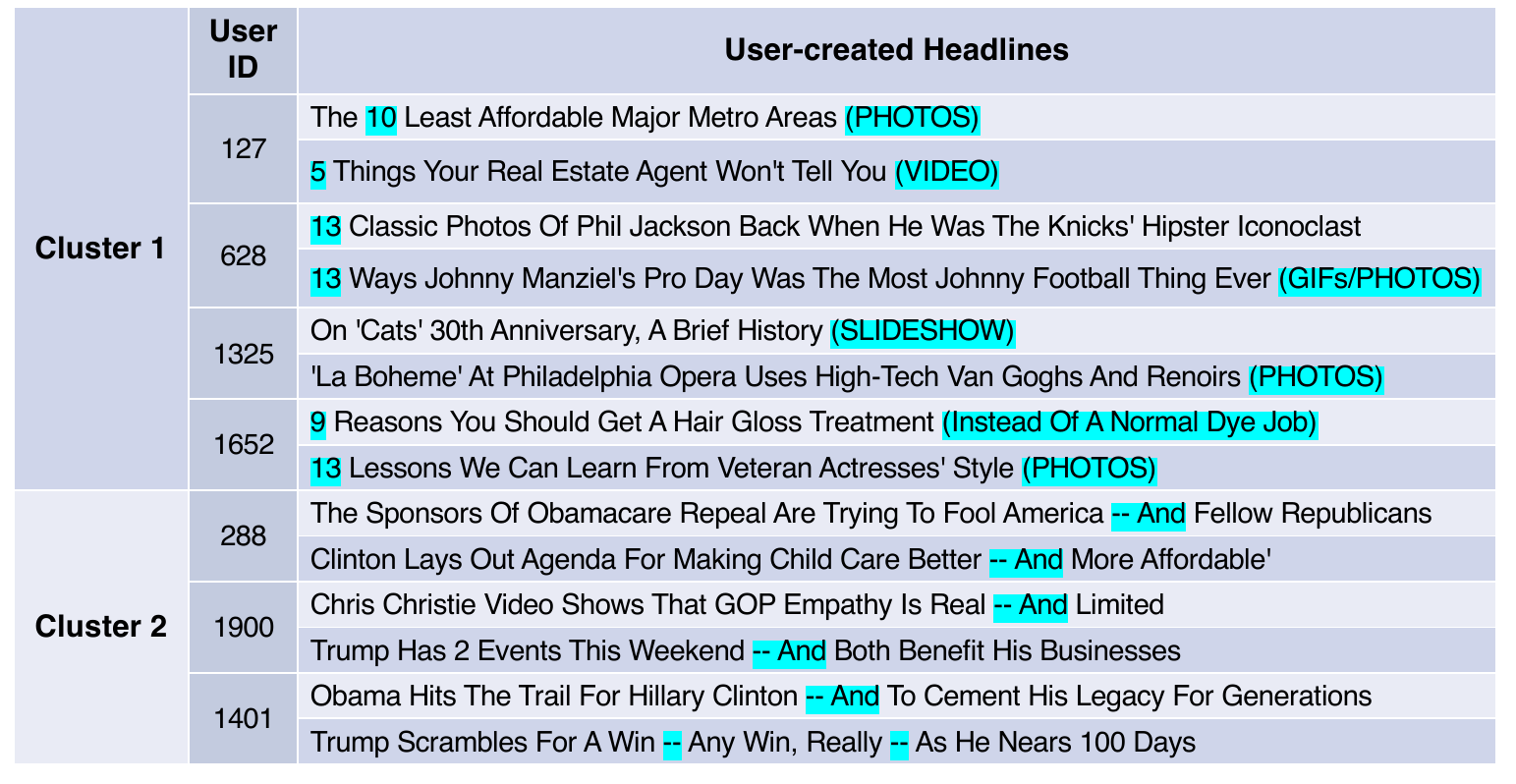} 
    \caption{Case study of clustering writing patterns in News Headline Generation task. The \blue{highlighted} tokens are the shared writing styles in cluster.}
    \label{fig:case_cluster}
\end{figure*}

\section{Scalability Bottlenecks of Baselines}
\label{app:efficiency}
As summarized in Table~\ref{tab:efficiency}, existing methods suffer from three critical scalability constraints:

\begin{itemize}
    \item \textbf{Training Time.} PETF demands user-specific adapter optimization with complexity $O(|P_u|)$, incurring significant costs for large user bases due to the heavy back-propagations. RAG is training-free, eliminating gradient-based training overhead but requiring $O(|P_u|)$ vectorization pre-processing.
    
    \item \textbf{Inference Latency.} In addition to the normal decoding latency of the language model, RAG suffers from $O(|P_u|)$ retrieval latency, which makes it inefficient for users with long histories. For PEFT, the process of loading these adapters can introduce overhead, particularly in scenarios requiring frequent updates or real-time interactions.
    
    \item \textbf{Storage Overhead.} RAG stores all historical interactions ($O(|P_u|)$ per user), scaling poorly for long-term usage. PETF maintains $O(rDL)$ storage for each user (typically 0.1\%-1\% of base model parameters), where $r$ is the rank of LoRA, $L$ is the number of layers and $D$ is the model hidden dimension.
\end{itemize}

\section{Datasets and Task Definition}
\begin{table}[t]
\begin{adjustbox}{max width=\linewidth}
\begin{tabular}{@{}ccccc@{}}
\toprule
                                   & \# User & Input Length & Output Length & \# History \\ \midrule
Abstract Generation      & 4560    & 33.82       & 144.28        & 120.30     \\
Topic Writing            & 2453    & 28.36       & 263.03        & 50.39      \\
Review Generation        & 1822    & 119.39      & 304.54        & 34.39      \\
News Headline Generation     & 2376    & 29.97       & 9.78          & 287.16     \\
Scholarly Title Generation & 2500    & 152.81      & 9.26          & 89.61      \\
Tweet Paraphrasing           & 1496    & 29.76       & 16.93         & 17.74      \\ \bottomrule
\end{tabular}
\end{adjustbox}
\caption{Datasets statistics. We report the number of users in test set, the averaged length of input $x$ and output $y$, and the averaged number of histories $|P_u|$.}
\label{tab:datasets}
\end{table}
This paper utilizes the LaMP benchmark and LongLaMP benchmark for evaluation. We only select generation tasks in the two benchmarks and the statistics are shown in Table~\ref{tab:datasets}. We show the input-output pair formats where the text in \{BRACES\} can be replaced with content specific to different users and queries:

\paragraph{LongLaMP: Abstract Generation} This task focuses on generating personalized abstracts for technical documents or articles based on the provided title and keywords.
\begin{center}
    \fcolorbox{black}{gray!20}{ \parbox{0.9\linewidth}{
    INPUT: Generate an abstract for the title "\{title\}" using the following items: "\{keywords\}" \\
    OUTPUT: \{abstract\}
    }}
\end{center}

\paragraph{LongLaMP: Review Generation} This task involves generating personalized product reviews that align with the user’s preferences, based on the product description and the score assigned to the product by the user.
\begin{center}
    \fcolorbox{black}{gray!20}{ \parbox{0.9\linewidth}{
    INPUT: Generate the review text written by a reviewer who has given an overall rating of \{rating\} for a product with description "\{description\}". The summary of the review text is "\{summary\}". \\
    OUTPUT: \{review\}
    }}
\end{center}

\paragraph{LongLaMP: Topic Writing} This task focuses on generating a personalized long-form Reddit post on a given topic from its summary written by user.
\begin{center}
    \fcolorbox{black}{gray!20}{ \parbox{0.9\linewidth}{
    INPUT: Generate the content for a Reddit post "\{summary\}". \\
    OUTPUT: \{post\}
    }}
\end{center}

\paragraph{LaMP: News Headline Generation} This task focuses on generating a personalized news headline for a given user-created article.
\begin{center}
    \fcolorbox{black}{gray!20}{ \parbox{0.9\linewidth}{
    INPUT: Generate a headline for the following article "\{article\}". \\
    OUTPUT: \{headline\}
    }}
\end{center}

\paragraph{LaMP: Scholarly Title Generation} This task requires language models to generate titles for an input abstract of an paper.
\begin{center}
    \fcolorbox{black}{gray!20}{ \parbox{0.9\linewidth}{
    INPUT: Generate a title for the following abstract of a paper "\{abstract\}". \\
    OUTPUT: \{title\}
    }}
\end{center}

\paragraph{LaMP: Tweet Paraphrasing} This task requires language models to generate a tweet in the style of a user given an input tweet.
\begin{center}
    \fcolorbox{black}{gray!20}{ \parbox{0.9\linewidth}{
    INPUT: Paraphrase the following tweet without any explanation before or after it "\{original tweet\}". \\
    OUTPUT: \{tweet\}
    }}
\end{center}

\end{document}